\documentclass[runningheads]{llncs}

\usepackage{eccv}

\usepackage{eccvabbrv}

\usepackage{graphicx}
\usepackage{booktabs}

\usepackage{multirow}
\usepackage{amsmath}
\usepackage{amssymb}
\usepackage{bbding}
\usepackage{colortbl}

\usepackage[accsupp]{axessibility}  

\usepackage{hyperref}

\usepackage{orcidlink}

\begin{document}

\title{See Through Their Minds: Learning Transferable Neural Representation from Cross-Subject fMRI} 

\titlerunning{See Through Their Minds}


\author{Yulong Liu\inst{1,2,3} \and
Yongqiang Ma\inst{1,2,3}\and
Guibo Zhu\inst{4,5}\and
Haodong Jing\inst{1,2,3}\and
Nanning Zheng\thanks{Corresponding Author} \inst{1,2,3}}

\authorrunning{Y.~Liu et al.}

\institute{National Key Laboratory of Human-Machine Hybrid Augmented Intelligence\and
National Engineering Research Center of Visual Information and Applications \and
Institute of Artificial Intelligence and Robotics, Xi’an Jiaotong University \and
 Institute of Automation, Chinese Academy of Sciences\and
University of Chinese Academy of Sciences\\
\email{\{lylhubxy, jinghd\}@stu.xjtu.edu.cn, \{musayq, nnzheng\}@mail.xjtu.edu.cn, gbzhu@nlpr.ia.ac.cn}
\vspace{-1em}}

   

\maketitle
\begin{abstract}
  Deciphering visual content from functional Magnetic Resonance Imaging (fMRI) helps illuminate the human vision system. However, the scarcity of fMRI data and noise hamper brain decoding model performance. Previous approaches primarily employ subject-specific models, sensitive to training sample size. In this paper, we explore a straightforward but overlooked solution to address data scarcity. We propose shallow subject-specific adapters to map cross-subject fMRI data into unified representations. Subsequently, a shared deeper decoding model decodes cross-subject features into the target feature space. During training, we leverage both visual and textual supervision for multi-modal brain decoding. Our model integrates a high-level perception decoding pipeline and a pixel-wise reconstruction pipeline guided by high-level perceptions, simulating bottom-up and top-down processes in neuroscience. Empirical experiments demonstrate robust neural representation learning across subjects for both pipelines. Moreover, merging high-level and low-level information improves both low-level and high-level reconstruction metrics. Additionally, we successfully transfer learned general knowledge to new subjects by training new adapters with limited training data. Compared to previous state-of-the-art methods, notably pre-training-based methods (Mind-Vis and fMRI-PTE), our approach achieves comparable or superior results across diverse tasks, showing promise as an alternative method for cross-subject fMRI data pre-training. Our code and pre-trained weights will be publicly released at \url{https://github.com/YulongBonjour/See_Through_Their_Minds}.
\end{abstract}

\section{Introduction}
\label{sec:intro}
\begin{figure*}[t]
\begin{center}
   \includegraphics[width=1.0\linewidth]{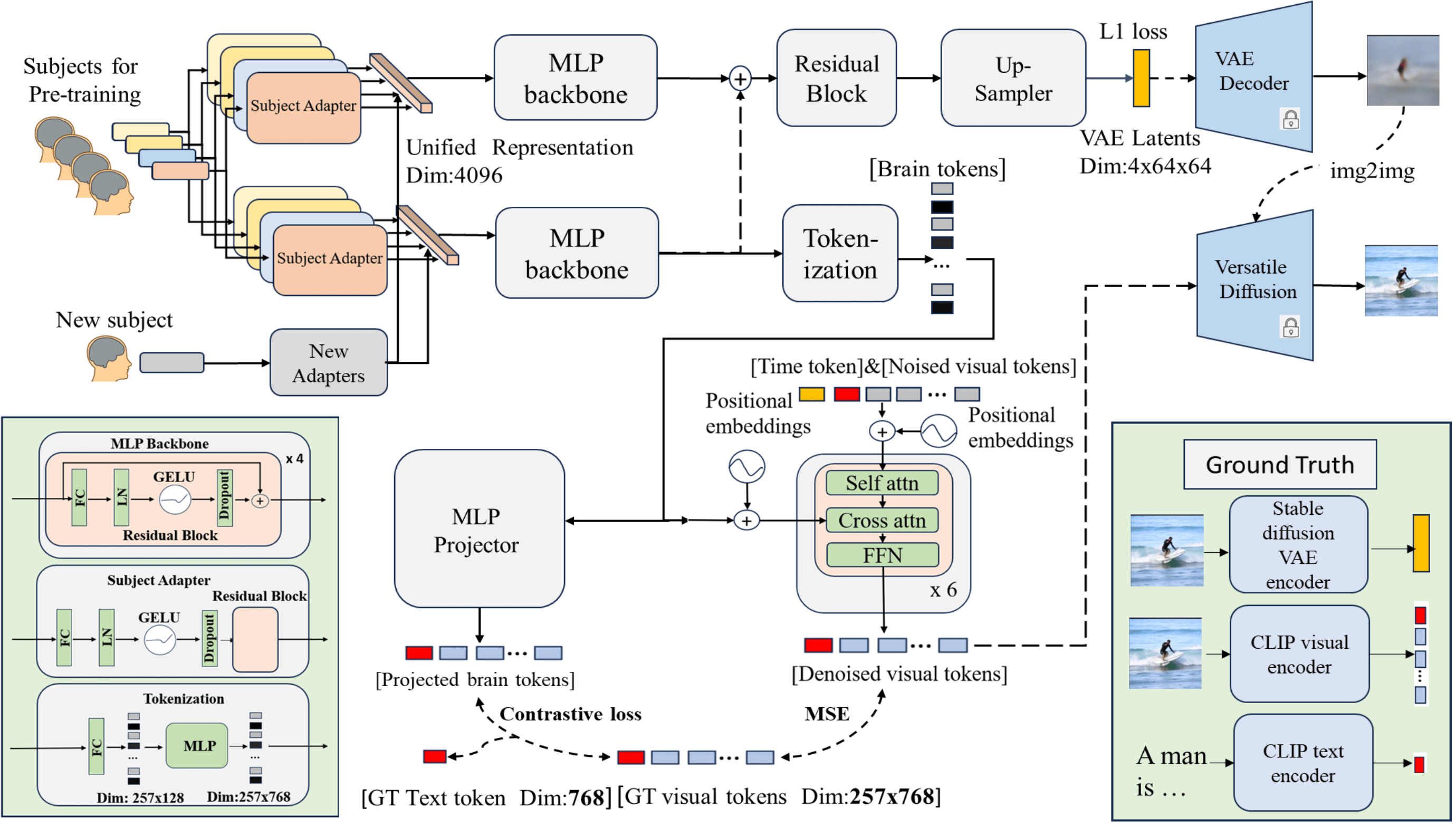}
\end{center}
   \caption{Overview of our STTM framework, which consists of a high-level perception decoding pipeline and a pixel-wise reconstruction pipeline(low-level pipeline). The two pipelines are trained sequentially. The pixel-wise reconstruction pipeline is guided by the high-level pipeline. The final reconstructions are generated in an img2img setting\cite{meng2021sdedit} using versatile diffusion model\cite{xu2023versatile}. Subject adapters are used to transform cross-subject fMRI data into a unified feature space for the two pipelines respectively. For new subjects, transfer learning can be conducted by training new adapters. The whole framework is inspired by the bottom-up and top-down processes in neuroscience.  \vspace{-1em}}
\label{fig:pipeline}
\end{figure*}
Vision decoding from brain activity enables the inference of mental states and cognitive processes, aiding neuroscience development and potentially advancing brain-inspired artificial intelligence and brain-computer interfaces.   To this end, functional Magnetic Resonance Imaging (fMRI) data are widely used as a non-invasive approach to capture brain activity.  Recent years have witnessed great progress in fMRI-based vision decoding with the rapid development of Deep Learning and generative models.
To name a few, Brain-diffuser\cite{ozcelik2023natural} adopts a two-stage reconstruction framework, where an fMRI pattern is first mapped into VAE latent embeddings to get a blurry reconstruction, then refining it using  predicted CLIP visual and textual embeddings to guide a pre-trained diffusion model. BrainCLIP\cite{liu2023brainclip} utilizes contrastive learning to align fMRI data with visual and textual modalities simultaneously, enabling versatile applications such as retrieval, classification, and reconstruction tasks. MindEye\cite{scotti2023reconstructing} enhances retrieval and reconstruction performance on the NSD\cite{allen2022massive} dataset by aligning fMRI data with fine-grained CLIP visual tokens through contrastive learning and diffusion priors\cite{ramesh2022hierarchical}.

While significant progress has been made, the subject-specific nature of most previous methods poses challenges to generalization. Firstly, fMRI data collection is costly and prone to blending with physiological noise, resulting in limited training samples with low signal-to-noise ratio per participant. Additionally, intrinsic inter-individual variability requires building models from scratch for each new subject, as every individual's brain has unique functional and anatomical structures. Consequently, subject-specific models may exhibit considerable variability in results. Despite achievements like those of MindEye on the NSD dataset, training models from scratch on small datasets remains prone to overfitting.

One strategy to somewhat alleviate data paucity is to pre-train foundation models on large-scale cross-subject data to acquire general knowledge inherent at the group level, which is gaining popularity in fMRI studies\cite{chen2023seeing,qian2023fmri,malkiel2022self,nguyen2020attend}, as it bypasses the need to train downstream models from scratch. Previous works have conducted some meaningful attempts in fMRI pre-training for vision decoding\cite{chen2023seeing,qian2023fmri}, where the fMRI data from different subjects are either split into patches similar to ViT\cite{dosovitskiy2020image} or transformed into unified 2D representations using anatomical information, then fMRI representations are learned using self-supervised tasks. Although these works provide new approaches to learning fMRI representation at the group level, individual variability is still the key challenge that hinders the final decoding performance. 


Recent studies indicate that coarse-grained response topographies exhibit high similarity across subjects, implying that individual idiosyncrasies are reflected in more nuanced response patterns\cite{chen2015reduced,gucclu2017increasingly,khosla2020shared}. This suggests the potential for decoding models to share representational spaces across subjects, mitigating challenges posed by limited per-subject data. Leveraging this insight, we propose a neural decoding model featuring  common decoding modules to capture shared response patterns, alongside subject-specific adapters that accommodate individual response biases. This approach enables the merging of data from multiple subjects viewing the same or different images to learn the general response patterns underlying different brains, while also capturing meaningful individual-level deviations. 


Our overall framework, referred to as \textbf{STTM}, is depicted in \cref{fig:pipeline}. Inspired by both bottom-up and top-down cognitive processes in the human brain \cite{katsuki2014bottom,miller1999straight}, we incorporate a high-level pipeline (\textbf{STTM-H}) to capture semantic-related perceptions and a low-level pipeline (\textbf{STTM-L}) focused on pixel-wise reconstruction to match the original images' low-level features (e.g., color, texture, spatial position). Both pipelines include subject-specific adapters for transforming cross-subject inputs, with the extracted features subsequently processed by shared decoding modules. The high-level pipeline aligns fMRI patterns with CLIP visual tokens using contrastive learning and a diffusion prior \cite{ramesh2022hierarchical}. Additionally, it aligns fMRI data with textual descriptions, enabling our model to be applied to multimodal brain decoding tasks. Our low-level pipeline involves mapping fMRI data onto the latent space of Stable Diffusion's variational autoencoder (VAE) \cite{rombach2022high} to obtain blurry reconstructions from the VAE's decoder. To enhance pixel-wise reconstruction, we propose utilizing high-level features from the MLP backbone of the high-level pipeline to guide the low-level pipeline. The final reconstructions are generated by combining outputs from the high-level and low-level pipelines in an img2img setting \cite{meng2021sdedit} using versatile diffusion\cite{xu2023versatile}. The over all design mirrors the bottom-up and top-down process in visual cortex.


In this study, we undertake two types of experiments: 1)  Model pre-training with data of four subjects (1, 2, 5, 7) from the Natural Scenes Dataset (NSD) \cite{allen2022massive}. Our results are compared with previous works on NSD. 2) Following pre-training, transfer learning is conducted on new subjects from the Generic Object Decoding (GOD) Dataset \cite{horikawa2017generic}, featuring much fewer training samples per subject and under a zero-shot setting.
Our contributions are summarized as follows:
\begin{itemize}
\item  To address the scarcity of fMRI data, we propose to use subject-adapters to transform cross-subject data into unified feature space and train shared decoding models to capture robust representations. This work contributes to the growing body of literature on pre-training and transfer learning with fMRI data.
\item We identify the importance of the interaction between high-level and low-level perceptions for reconstruction performance. To leverage this interaction, we propose utilizing high-level perceptions to guide pixel-wise reconstruction. Our overall framework, employing an img2img setting, simulates the bottom-up and top-down processes observed in neuroscience and obtains reasonable performance gains. 
\item We open source a versatile brain decoding model with good transferability and high performance on a wide range of tasks, which may facilitate future multi-model brain decoding research.
\end{itemize}

\section{Related Work}
\subsubsection{Visual Stimulus Decoding from FMRI.}
Deciphering visual information from human brain activity has long been a pursuit in neuroscience. Due to challenges such as low signal-to-noise ratio and limited fMRI samples, early studies primarily employed linear models to map fMRI data onto an intermediate feature space for decoding basic visual attributes like spatial position\cite{thirion2006inverse}, orientation\cite{haynes2005predicting,kamitani2005decoding}, and image categories\cite{cox2003functional,haxby2001distributed}. As deep learning and generative models rapidly advanced, researchers explored reconstructing visual stimuli using CNNs\cite{shen2019deep,shen2019end,beliy2019voxels}, GANs\cite{ren2021reconstructing,ozcelik2022reconstruction,lin2022mind,mozafari2020reconstructing}, and diffusion models\cite{liu2023brainclip,ozcelik2023natural,scotti2023reconstructing}, resulting in increasingly semantically plausible and faithful reconstructions. The emergence of models like CLIP\cite{radford2021learning} and related ones such as Stable diffusion\cite{rombach2022high} and versatile diffusion\cite{xu2023versatile} further fueled exploration in decoding approaches. State-of-the-art methods have recently integrated contrastive learning\cite{oord2019representation} between fMRI data and CLIP model features to improve neural representations, followed by reconstruction using the Diffusion model or GANs. Notably, Mind-Reader\cite{lin2022mind} and BrainCLIP\cite{liu2023brainclip}  utilized the CLS token from CLIP's visual and textual encoder as global supervision, while MindEye\cite{scotti2023reconstructing} leveraged all 257 tokens from CLIP's visual encoder's last hidden layer, showing the benefits of fine-grained supervision for retrieval and reconstruction tasks. In this paper, we propose a fusion of global visual-linguistic contrastive learning and fine-grained visual contrastive learning to enhance multi-modal brain decoding. We also emphasize the importance of the interaction between bottom-up and top-down processes for stimulus reconstruction.

\subsubsection{FMRI Foundation Models for Visual Decoding.} 
The pre-training-finetuning paradigm has demonstrated significant success across various domains\cite{devlin2018bert,radford2021learning,he2022masked,guwukong,liutaisu}. While most previous brain decoding methods have traditionally employed subject-specific pipelines, recent efforts have emerged to develop fMRI foundation models by pretraining decoding models with large-scale fMRI data collected from diverse subjects, aiming to capture generalizable neural representations across different brains\cite{chen2023seeing,qian2023fmri,malkiel2022self}. Chen et al.\cite{chen2023seeing} divided fMRI series into patches and transformed them into embeddings, utilizing mask brain modeling (MBM) to learn fMRI representations. Malkiel et al.\cite{malkiel2022self} applied self-supervised techniques, including MBM, to learn representations of audio-evoked fMRI data.
More recently, Qian et al.\cite{qian2023fmri} transformed fMRI signals into unified 2D representations using anatomical information, which can maintain consistency across individuals while preserving distinct brain activity patterns and are then used to train foundation models. It is worth noting that these pre-trained models still require additional fine-tuning on an individual basis to accommodate the intricate biological nuances governing visual stimuli generation.  In our work, instead of employing the same network to process fMRI data from different subjects, we propose training shallow subject-specific adapters alongside a shared deep decoding network. And transfer learning can be conducted by training new adapters for new subjects.

\section{Method}

Drawing inspiration from the bottom-up and top-down processes in the human brain\cite{katsuki2014bottom,miller1999straight}, our method adopts an img2img setting which consists of a high-level perception decoding pipeline and a low-level pixel-wise reconstruction pipeline guided by high-level perceptions. Further details will be provided in the subsequent subsections.

\subsection{Cross-Subject High-Level Perceptions Decoding}
Our high-level pipeline is designed to capture semantic perceptions underlying the fMRI data, aligning cross-subject fMRI data with visual and textual modalities. This versatility allows it to be applied to tasks such as fMRI-to-image retrieval, fMRI-to-text retrieval, zero-shot classification, and fMRI-to-image generation.
\subsubsection{High-Level Model Architecture}
Our pipeline aims to translate fMRI data into CLIP embedding space.  It consists of several shallow subject-specific adapters, a shared MLP backbone with 4 residual blocks, a tokenization module, a diffusion prior module, and an MLP projector. The subject adapters are formed by a linear projection followed by a residual block, accommodating individual variability and aligning fMRI data across subjects into a unified feature space. The shared residual MLP backbone further refines these features into a higher-level feature space. Subsequently, the tokenization module transforms the extracted features into 257 fine-grained tokens, corresponding to the 257 visual tokens from CLIP's visual encoder's last hidden layer. Unlike MindEye, which directly projects hidden states from the shared backbone into tokens with a dimension of 257x768 via linear projection, our tokenization module first maps the hidden states to a low-dimension space(257x128) and then uses a single-hidden-layer MLP to ascend the dimension to 257x768, which significantly reduces the number of parameters while maintains high-performance(See \cref{nsd retrival}).
 The tokens are then processed by an MLP projector and a diffusion prior module in parallel. The model is trained end-to-end, with the prior receiving an MSE loss and the projector receiving a bidirectional contrastive loss. Projector outputs are for retrieval tasks, and diffusion prior outputs are used to guide image generation of a pre-trained versatile diffusion model\cite{xu2023versatile}. Unless specified otherwise, the CLIP model used in this work is CLIP ViT\slash L-14.

\subsubsection{Global Visual-Linguistic Contrastive Learning $\&$ Fine-Grained Visual Contrastive Learning}
Previous studies have demonstrated the efficacy of contrastive learning in producing robust fMRI representations. Notably, BrainCLIP\cite{liu2023brainclip} combines global embeddings from both CLIP's visual and textual encoders as target features, showing improved reconstruction performance compared to conditions with only visual supervision. MindEye\cite{scotti2023reconstructing} utilizes all 257 tokens from CLIP's visual encoder's last layer, showcasing the benefits of fine-grained supervision for both image retrieval and reconstruction performance.
To further enhance expressive representations that bridge the brain with visual and textual modalities for potential multi-modal brain decoding, we propose combining global visual-linguistic contrastive learning (GVLC) and fine-grained visual contrastive learning (FVC). Specifically, FVC involves contrasting the flattened and L2-normalized 257 CLIP visual tokens ($V_f$) with the 257 projected Brain tokens ($B_f$) from the MLP projector,
\begin{equation}\label{eq1}
L_{FVC}=Contrast(V_f,B_f).
\end{equation}
And the GVLC is applied to the CLS token($B_{CLS}$) of the Brain tokens by combine the supervision from CLIP visual and textual CLS token($V_{CLS}$ and $T_{CLS}$), i.e.,
\begin{equation}\label{eq2}
L_{GVLC}=\frac{1}{2}[Contrast(V_{CLS},B_{CLS})+Contrast(T_{CLS},B_{CLS})].
\end{equation}
Two kinds of contrastive loss are used in this work. For the first 35\% percent of the total epochs, we use BiMixCo loss proposed in \cite{scotti2023reconstructing}, which uses Mixup technique to train models on synthetic fMRI data created through convex combinations of two fMRI-stimulus pairs, aiming to alleviate the data scarcity for a single subject. For the rest epochs, SoftCLIP loss \cite{scotti2023reconstructing} is used as the contrastive loss, which is inspired by knowledge distillation\cite{hinton2015distilling} and uses the batch-wise visual CLIP embedding similarity instead of one-hot labels as the target label. Both the two kinds of contrastive loss are bidirectional(See Appendix \cref{loss} for details).

\subsubsection{Efficient Diffusion Prior Learning}

Contrastive learning yields disjoint fMRI embeddings, known as the "Modality Gap"\cite{liang2022mind}. To reconstruct images, we train a diffusion prior proposed in DALL-E2\cite{ramesh2022hierarchical} to produce aligned CLIP embeddings from the outputs of the MLP backbone. These embeddings can serve as inputs to any pre-trained image generation model accepting CLIP image embeddings. MindEye applied this technique to map fMRI to CLIP image embeddings, predicting denoised CLIP tokens based on brain tokens and noise tokens. However, MindEye's encoder-only transformer architecture demands significant GPU memory for long sequences. In our work, we employ a 6-layer transformer decoder architecture for the diffusion prior to mitigate GPU memory usage.
The diffusion prior receives three token types: brain tokens, noisy CLIP visual tokens, and a time token. Brain tokens serve as keys and values for cross-attention modules in the transformer decoder after adding positional embeddings. The time token and noisy CLIP visual tokens are concatenated as queries, with positional embeddings added to the latter. The diffusion prior module are trained to output denoised CLIP tokens, utilizing the same diffusion loss as Ramesh et al.\cite{ramesh2022hierarchical}.
Inspired by the success of masked autoencoder\cite{he2022masked}, we propose training the diffusion prior with only a small part(e.g., 35\% ) of the predicted Brain tokens. This approach offers dual benefits: it further reduces GPU memory usage during training and enhances the expressiveness of each brain token.

Our total end-to-end loss  for the high-level pipeline is defined as:
\begin{equation}\label{eq3}
L_{high}=\alpha(L_{FVC}+\beta L_{GVLC})+(1-\alpha)L_{prior}.
\end{equation}

In our experiments, we set $\beta$ to 0.4 and employ random weighting\cite{lin2021reasonable} between the contrastive losses and the diffusion prior loss.

\subsection{Cross-Subject Pixel-Wise Reconstruction with High-Level Perception Guidance}
The human brain processes visual content through two distinct information pathways: bottom-up and top-down \cite{katsuki2014bottom,miller1999straight}. In bottom-up pathways, local features integrate to shape the brain's understanding of global information, while in top-down pathways, the brain adjusts its perception of low-level features based on semantic understanding. These pathways act as complementary forces, often collaborating to help us comprehend complex stimuli.
Given that CLIP visual and textual tokens primarily encode high-level semantic information rather than low-level details such as texture and boundary, it becomes imperative to incorporate a low-level reconstruction pipeline to preserve pixel-level information from fMRI data. To address this need, we propose a pipeline that integrates both bottom-up processing and top-down feedback mechanisms.

As depicted in \cref{fig:pipeline}, our low-level pipeline consists of shallow subject-specific adapters followed by a shared residual MLP backbone. The outputs from this backbone receive semantic information processed by the high-level pipeline backbone, with the two sets of information added together. Subsequently, the merged features undergo further processing by a residual block before being up-sampled to the latent space of Stable-diffusion's VAE by CNN.
During training, we employ L1 loss as the loss function. To fully utilize both high-level and bottom-up information, we introduce a mechanism where, in 30\% of training steps, the semantic feedback is substituted with a learnable embedding vector. Additionally, in another 25\% of training steps, the outputs from the low-level pipeline backbone are replaced with another learnable embedding vector. After training, blurry reconstructions can be generated by decoding the predicted latent embeddings with the VAE decoder. These reconstructions serve as the initial states for the versatile diffusion model.

\begin{table}[t]
    \centering
      \caption{Comparison with several previous works on retrieval tasks. For the image and brain retrieval tasks, the candidate pool size is 300, and top-1 accuracy is reported. For the text retrieval task, the candidate pool size is 982,  and top-5 accuracy is reported since some test samples have similar captions. In comparison with MindEye, our models demonstrate better parameter efficiency and versatility.}
       \label{nsd retrival}
    \begin{tabular}{cccccc}
        \toprule
        \multirow{2}*{Methods}& \multirow{2}*{Multi-subject}& \multirow{2}*{ Parameters}&\multicolumn{3}{c}{Retrieval tasks}\\
        \cmidrule(r){4-6}
        &&&Image@1$\uparrow$& Text@5$\uparrow$ &Brain@1$\uparrow$\\
        \midrule
       Mind Reader\cite{lin2022mind}&\XSolidBrush&2.34M&11.0\%&-&49.0\% \\
       Brain-diffuser\cite{ozcelik2023natural}&\XSolidBrush&3B&21.1\%&-&30.3\%\\
       BrainCLIP-VAE\cite{liu2023brainclip}&\XSolidBrush&18.6M&40.65\% &31.1\%&-\\
       MindEye(High-level)\cite{scotti2023reconstructing}&\XSolidBrush&1B&93.6\%&-&90.1\% \\
       STTM-H(Ours,w/o GVLC) &\Checkmark&568M&\bf{93.6\%}& 10.3\%&94.8\% \\
       \rowcolor{gray!20}
       STTM-H(Ours)&\Checkmark&568M&92.8\% &\bf{41.3\%} &\bf{94.9\%}\\
      \bottomrule
      \end{tabular}
\end{table}
\vspace{-1em}

\subsection{Transfer learning for new subjects}
Deciphering brain activity of new subjects using models trained for other individuals has been a longstanding goal in neuroscience. Here, we propose an adapter-based approach to alleviate this challenge. In our brain decoding scenario, we hope the decoding performance for new subjects who may have limited fMRI data can benefit from the general knowledge learned from abundant cross-subject fMRI patterns. To achieve this, we freeze the pre-trained shared decoding modules and train a shallow adapter for each of the two pipelines, aligning the fMRI data of the new subject with the unified feature space. The training objective is the same as the pre-training stage. Once the subject adapter is trained, we freeze both the adapter and the MLP backbone. Subsequently, we fine-tune the remaining modules to further align the fMRI patterns with the target domain.

\section{Experimental Results}
\subsection{Datasets and Setting}

 \subsubsection{Natural Scenes Dataset (NSD)} The NSD dataset\cite{allen2022massive}, sourced from 8 subjects viewing images from the COCO dataset\cite{lin2014microsoft}, is currently the largest neural imaging dataset for data-driven brain decoding. Our study uses the data of subjects 1, 2, 5, and 7 from NSD. Each subject's training set includes 8859 image stimuli and 24980 fMRI trials, while the test set comprises 982 image stimuli and 2770 fMRI trials, with stimuli differing across subjects in the training set but shared in the test set. Responses for images with multiple trials are averaged. Utilizing the NSDGeneral ROI mask at 1.8 mm resolution, we derived ROIs for the 4 subjects, encompassing 15724, 14278, 13039, and 12682 voxels, spanning visual areas from early to higher visual cortex. Corresponding captions can be extracted from the COCO dataset.
  \begin{figure*}[t]
\begin{center}
   \includegraphics[width=1.0\linewidth]{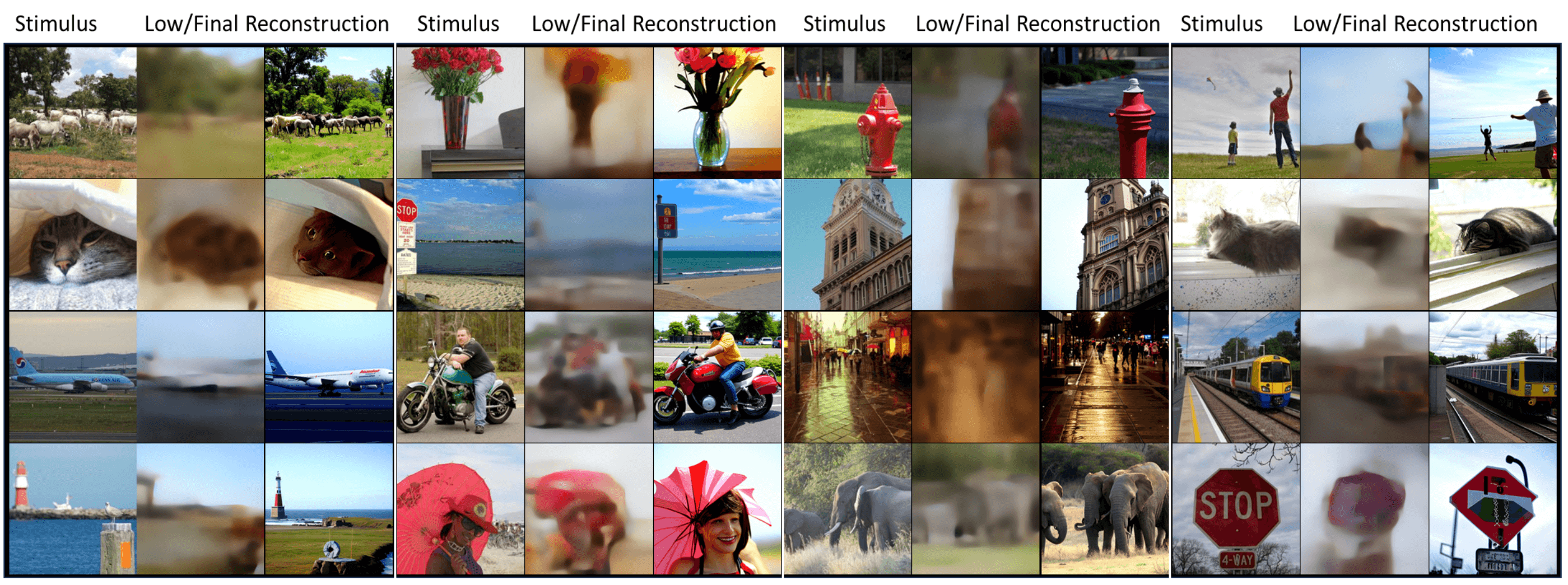}
\end{center}
   \caption{Reconstruction examples for subject 1 in the NSD dataset. The "Low reconstructions" are from the low-level pipeline and the "final reconstructions" are obtained in the img2img \cite{meng2021sdedit} setting.  \vspace{-1em}}

\label{fig:nsd_examples}
\end{figure*}

\subsubsection{Generic Object Decoding(GOD) Dataset}

   The GOD dataset, curated by Horikawa and Kamitani\cite{horikawa2017generic}, comprises fMRI recordings of five healthy subjects viewing images from ImageNet\cite{deng2009imagenet}. It includes 1250 images from 200 ImageNet categories, with 1200 training images from 150 categories and 50 test images from the remaining categories. Training and test stimuli were presented 1 and 35 times to the subjects, respectively, resulting in 1200 and 1750 fMRI instances. We utilize preprocessed regions of interest (ROIs)\footnote{The preprocessed data and demo code are available at \url{http://brainliner.jp/data/brainliner/Generic_Object_Decoding}} covering voxels from early to higher visual areas. FMRI data from different trials for each test image are averaged. Additionally, we acquire caption annotations from the GOD-Cap dataset\cite{liu2023brainclip}, offering two textual descriptions for each visual stimulus.
\begin{table*}[t]
 \caption{Quantitative comparison of STTM reconstruction performance against other models.
All results are averaged across the same 4 participants (see Appendix \cref{nsd_individual} for individual subject models), except Lin et al.\cite{lin2022mind} which only analyzed Subject 1. The middle section and bottom section do ablation studies for the low-level and high-level pipelines respectively, and also compare our method with MindEye. We use the same evaluation metrics(See Appendix \cref{metrics} for metric details) and the same image preprocessing as Brain-diffuser\cite{ozcelik2023natural} and MindEye\cite{scotti2023reconstructing}. Bold indicates best performance within sections.
}
 \resizebox{\linewidth}{!}{
    \centering
    \begin{tabular}{ccccccccc}
        \toprule
        \multirow{2}*{Methods}&\multicolumn{4}{c}{Low- Level}&\multicolumn{4}{c}{High-Level} \\
        \cmidrule(r){2-5} \cmidrule(r){6-9}\\
        &PixCorr$\uparrow$&SSIM$\uparrow$&Alex(2)$\uparrow$&Alex(5)$\uparrow$&Incep$\uparrow$& CLIP$\uparrow$&Eff$\downarrow$&SwAV$\downarrow$\\
        \midrule
      Mind Reader\cite{lin2022mind} & -&-&-&-&78.2\%&-&-&-\\
      Takagi...\cite{takagi2022high}&-&-&83.0\%&83.0\%&76.0\%&77.0\%&-&-\\
      Gu et al.\cite{gu2022decoding}&.150&.325&-&-&-&-&.862&.465\\
      Brain-diffuser\cite{ozcelik2023natural}&.254&\bf{.356}&94.2\%&96.2\%&87.2\%&91.5\%&.775&.423\\
      BrainCLIP\cite{liu2023brainclip}&-&-&-&-&86.7\%&94.8\%&-&-\\
      fMRI-PTE (MG)\cite{qian2023fmri}& .131 &.112 &78.13\% &88.59\% &84.09\% &82.26\% &.837 &.434\\
      DREAM\cite{xia2024dream}&.288&.338&93.9\%&96.7\%&93.7\%&94.1\%&.645&.418\\
MindEye\cite{scotti2023reconstructing}&.309&.323&94.7\%&97.8\%&93.8\%&94.1\%&.645&.367\\
MindEye-BOI\cite{kneeland2023brain}&.259&.329&93.9\%&97.7\%&93.9\%&93.9\%&.645&.367\\
       \rowcolor{gray!20}
       STTM(Ours)&\bf{.333}&.334&\bf{95.7\%}&\bf{98.5\%}&\bf{95.8\%}&\bf{95.7\%}&\bf{.611}&\bf{.338}\\
      \midrule
      MindEye(Low-level)\cite{scotti2023reconstructing}&.360&.479&78.1\%&74.8\%&58.7\%&59.2\%&1.0&.663\\
      STTM-L(Ours,w/o guidance)&.372&.488&79.6\%&79.6\%&63.6\%&63.0\%&.985&\bf{.643}\\
       \rowcolor{gray!20}
       STTM-L(Ours,with guidance)&\bf{.383}&\bf{.488}&\bf{83.3\%}&\bf{86.0\%}&\bf{68.2\%}&\bf{67.1}\%&\bf{.968}&.647\\
      
      \midrule
      MindEye(High-level)\cite{scotti2023reconstructing}&.194&\bf{.308}&\bf{91.7\%}&97.4\%&93.6\%&94.2\%&.645&.369\\
      STTM-H(Ours,w/o GVLC)&.201&.276&91.4\%&97.8\%&95.3\%&95.1\%&.622&.352\\
       \rowcolor{gray!20}
       STTM-H(Ours)&\bf{.209}&.276&91.5\%&\bf{97.8\%}&\bf{95.4\%}&\bf{95.6\%}&\bf{.612}&\bf{.344}\\
      \bottomrule
      \end{tabular}}
       \label{nsd_gen_compare}
       \vspace{-1em}
\end{table*}
\subsubsection{Implementation Details}

Our models are trained and tested on 8 Hygon DCUs with 16GB HBM2 memory. With the data of the 4 subjects from NSD, We pre-train the high-level pipeline for 280 epochs and the low-level pipeline for 540 epochs with a global batch size of 192. For the High-level pipeline on the GOD dataset, we first train a new subject adapter for 4500 epochs with a global batch size of 880 for each new subject, while keeping the pre-trained parts frozen. Then, we freeze the adapter and the MLP backbone, and fine-tune the rest  parts of the model for 800,400,400,400,800 epochs with a global batch size of 600 for subjects 1,2,3,4,5 respectively. Similarly, for the low-level pipeline on the GOD dataset, we train a new subject adapter for 5000 epochs with a global batch size of 192 for each new subject, with the pre-trained parts frozen. Then, we freeze the adapter and MLP backbone, and fine-tune the rest parts for 800 epochs. We use AdamW\cite{loshchilov2018decoupled} for optimization with $\beta_1=0.9$, $\beta_2=0.999$, and $\epsilon=10^{-8}$. Additionally, we employ the OneCircle learning rate schedule\cite{smith2019super} with a maximum learning rate of 0.0005. For reconstruction evaluation metrics, we use the implementation of MindEye. More details can be found in our code. 

\subsection{Brain-Image/Text Retrieval on NSD}
For image retrieval or text retrieval, the goal is to match the correct image or text with a given fMRI pattern among multiple candidates. In image retrieval, we calculate cosine similarity between the flattened and normalized 257 tokens of a brain sample and each of 300 randomly selected image candidates from the test set. This process is repeated for each of the 982 brain samples in the test set, and the overall accuracy is averaged across 30 iterations to accommodate batch sampling variability. In text retrieval, the candidate pool for each fMRI pattern comprises all 982 image captions, and cosine similarity is computed between the CLS token of the fMRI pattern and those of the candidate captions. For brain retrieval, the procedure mirrors image retrieval, but image and brain samples are swapped to find the corresponding brain sample for a given image among 300 brain samples.

In \cref{nsd retrival}, we compare our method with previous works across retrieval tasks, reporting top-1 accuracy for image and brain retrieval and top-5 accuracy for text retrieval due to similar captions for some test images. Our method demonstrates strong generalization across all retrieval tasks, achieving superior performance in text retrieval and brain imaging retrieval. Compared to MindEye, the prior state-of-the-art method, our method exhibits better parameter efficiency and versatility.

\subsection{FMRI-to-Image Reconstruction on NSD}
We generate 16 CLIP image embeddings with diffusion prior for each test brain sample, then pass them through the image variations pipeline of Versatile Diffusion, starting the denoising process with the blurry reconstruction from our low-level pipeline. This process comprises 20 timesteps using UniPCMultistep noise scheduling\cite{zhao2024unipc},  yielding 16 reconstructions per sample. Finally, we select the best reconstruction using our retrieval branch.

\begin{table}[t]
    \centering
     \caption{\textbf{Zero-shot visual stimulus classification on GOD dataset}. The test set contains 50 categories that have no overlapping with the training set. The results for CADA-VAE, MVAE, MMVAE, MoPoE-VAE, and BraVL are taken from \protect\cite{du2023decoding}. The results for LEA\cite{qian2024lea} and BrainCLIP\cite{liu2023brainclip} are obtained from respective papers.V\&T means that the model is trained with visual and textual features while V(T) means that the model is trained with only visual(textual) features. All results are averaged across 5 subjects.}
    \begin{tabular}{ccccc}
        \toprule
        \multirow{2}*{Methods}& \multirow{2}*{Modality}&\multirow{2}*{Prompt}&\multicolumn{2}{c}{Average}\\
        \cmidrule(r){4-5}
        &&&top-1 &top-5\\
        \midrule
        CADA-VAE \cite{schonfeld2019generalized}&V\&T &- &10.0\%&40.4\%\\ 
        MVAE \cite{wu2018multimodal} &V\&T &- &10.0\%&39.6\%\\   
        MMVAE \cite{shi2019variational} &V\&T &- &11.7\%&43.3\%\\
        MoPoE-VAE \cite{sutter2021generalized}&V\&T &-&12.9&51.8\\ 
        BraVL \cite{du2023decoding} &V\&T &- &14.0\%&53.1\%\\  
        LEA\cite{qian2024lea}&V&-&13.6\%&- \\
        BrainCLIP-VAE\cite{liu2023brainclip} &V\&T &Text &18.4\%&51.2\%\\
        BrainCLIP-VAE \cite{liu2023brainclip}&V\&T &CoOp &18.0\%&59.5\%\\
         \rowcolor{gray!20}
        STTM-H(Ours)&V\&T&Text&\bf{23.2}\%&\bf{62.0}\%\\
        
      \bottomrule
      \end{tabular}
      \vspace{-1em}
       \label{zero_classfication}
\end{table}

Reconstruction examples for the NSD dataset are shown in \cref{fig:nsd_examples}. The low-level reconstructions  effectively capture details such as position and color distribution, while the final reconstructions enhance the semantic recognizability of the initial blurry low-level reconstructions. In \cref{nsd_gen_compare}, we quantitatively compare our method with recent works on the NSD dataset. Our full method outperforms previous approaches by significant margins across low-level and high-level metrics, except for SSIM. Specifically, STTM-L significantly outperforms MindEye(Low-level) on all metrics, especially on high-level metrics, indicating superior semantic recognizability of our low-level reconstructions. STTM-H also outperforms MindEye(High-level) by a considerable margin on all high-level metrics, demonstrating better capture of semantic information. 

Additionally, compared to fMRI-PTE\cite{qian2023fmri}, which was pre-trained with large-scale unsupervised fMRI data from 1000 subjects before training on the NSD dataset, our method exhibits clear advantages in results. This suggests greater efficiency of our method in learning general knowledge from cross-subject data. Moreover, the img2img strength used for the results in \cref{nsd_gen_compare} is 0.3, and we have tested with other values, obtaining robust performances (see Appendix \cref{img2img}). 

\begin{figure*}[t]
\begin{center}
   \includegraphics[width=1.0\linewidth]{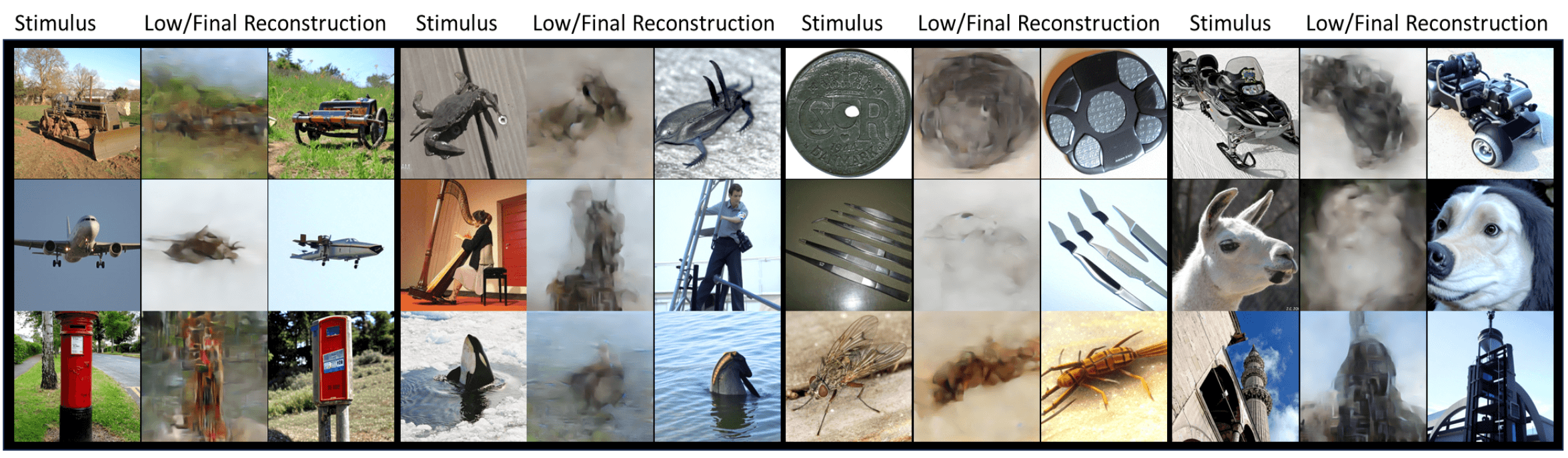}
\end{center}
   \caption{Reconstruction examples for subject 3 in GOD  \vspace{-1em}}
\label{fig:GOD_examples}
\end{figure*}
\subsection{Transfer Learning on GOD Dataset}
In this section, we aim to transfer the knowledge acquired from the NSD dataset to the GOD dataset, which is considerably smaller and in a zero-shot setting. We assess the transfer learning outcomes through two tasks: the zero-shot classification task and the image reconstruction task. 

Additionally, we conducted tests to assess whether our decoders, originally trained on brain activity induced by visual stimuli, possess the capability to generalize to decode imagery-induced brain activity(See Appendix \cref{imagery}).

\subsubsection{Zero-Shot Classification on GOD by Prompting}
Utilizing CLIP's well-aligned embedding space, STTM enables zero-shot visual stimulus classification. This is achieved by comparing the embeddings of the CLS token of Brain tokens with the classification weights synthesized by CLIP's text encoder, which takes as input textual prompts specifying classes of interest. These prompts can be manually designed, such as "a photo of a [CLASS]," where the class token is replaced by the specific class name, and other tokens provide context for the class name. We employ the same text prompts as BrainCLIP. The results are shown in \cref{zero_classfication}. Our method obtains an average top-1 accuracy of 23.2\% and an average top-5 accuracy of 62.0\%, which are considerably better than previous works.


\subsubsection{Image Reconstruction on GOD}
In \cref{fig:GOD_examples}, we present visualizations of low-level and final reconstructions for the GOD dataset. In \cref{GOD_gen_compare}, we provide our average results across five subjects and compare our method with recent works, focusing on subject 3, as only results for this subject are available for Mind-Vis\cite{chen2023seeing} and CMVDM\cite{zeng2023controllable}.  As we can see, due to the limited training samples and low signal-to-noise ratio, none of the models outperform the others across all metrics. Our low-level pipeline achieves the best performance on PixCorr and SSIM, two pixel-wise metrics. Additionally, our final reconstruction compares favorably with previous state-of-the-art works. Notably, Mind-Vis is also a pertaining-based decoding method, using masked signal modeling for pre-training and then fine-tuning on downstream datasets. Our method outperforms Mind-Vis on most metrics, suggesting it is a promising alternative for pre-training fMRI models.
 
 \begin{table*}[t]
    \centering
     \caption{Evaluation of image reconstruction on GOD. Our results are obtained by transfer learning.  Since recent state-of-the-art methods like Mind-Vis\cite{chen2023seeing} and CMVDM\cite{zeng2023controllable} only report their results on subject 3, We present our average performances at the top and compare against other methods specifically on subject 3 at the bottom. The results for Mind-Vis and CMVDM are calculated based on their reported reconstructions, while the results for IC-GAN are obtained by rerunning its model with provided weights.}
    \begin{tabular}{ccccccccc}
        \toprule
        \multirow{2}*{Methods}&\multicolumn{4}{c}{Low- Level}&\multicolumn{4}{c}{High-Level} \\
        \cmidrule(r){2-5} \cmidrule(r){6-9}\\
        &PixCorr$\uparrow$&SSIM$\uparrow$&Alex(2)$\uparrow$&Alex(5)$\uparrow$&Incep$\uparrow$& CLIP$\uparrow$&Eff$\downarrow$&SwAV$\downarrow$\\
        \midrule
      STTM-H(Ours)&.143&.341&85.2\%&91.2\%&77.8\%&82.9\%&\bf{.871}&.534\\
    STTM-L(Ours)&\bf{.274}&\bf{.500}&87.9\%&91.0\%&64.2\%&58.2\%&.956&.711\\
        STTM(Ours)&.202&.366&\bf{90.3\%}&\bf{93.4\%}&\bf{80.0\%}&\bf{84.1}\%&.879&\bf{.526}\\
        \midrule
        IC-GAN(Sub3) \cite{ozcelik2022reconstruction}&.195& .386&88.1\%&95.3\%&\bf{85.8\%}&84.0\%&.855&.486\\
        MinD-Vis(Sub3)\cite{chen2023seeing}& .119&.390&82.8\%&93.8\%&81.0\%&82.6\%&.833&.491\\
        CMVDM(Sub3)\cite{zeng2023controllable}&.279 &.454&88.4\%&93.9\%&81.6\%&82.0\%&\bf{.810}&\bf{.485}\\
        
        STTM-H(Ours,Sub3)&.133&.328&87.9\%&93.3\%&79.7\%&86.4\%&.851&.521\\
         \rowcolor{gray!20}
        STTM-L(Ours,Sub3)&\bf{.322}&\bf{.501}&90.3\%&92.7\%&66.8\%&58.2\%&.954&.704\\
         \rowcolor{gray!20}
         STTM(Ours,Sub3)&.253&.367&\bf{92.0\%}&\bf{95.6\%}&81.3\%&\bf{87.0\%}&.870&.505\\
      \bottomrule
      \end{tabular}
       \label{GOD_gen_compare}
        \vspace{-1em}
\end{table*}

\begin{table}[h]
 \caption{Further validation of the effects of cross-subject training on subject 1 of NSD. There are 3 types of conditions. Type I means both STTM-L and STTM-H are trained with single-subject data. Type II means the STTM-L is trained with single-subject data but the high-level guidance is from STTM-H trained with cross-subject data. Type III means both STTM-H and STTM-L are trained with cross-subject data.
}
 \resizebox{\linewidth}{!}{
    \centering
    \begin{tabular}{ccccccccc}
        \toprule
        \multirow{2}*{Methods}&\multicolumn{4}{c}{Low- Level}&\multicolumn{4}{c}{High-Level} \\
        \cmidrule(r){2-5} \cmidrule(r){6-9}\\
        &PixCorr$\uparrow$&SSIM$\uparrow$&Alex(2)$\uparrow$&Alex(5)$\uparrow$&Incep$\uparrow$& CLIP$\uparrow$&Eff$\downarrow$&SwAV$\downarrow$\\
        \midrule
      STTM-L(Sub1, \uppercase\expandafter{\romannumeral 1}) &.417&.487&87.9\%&87.8\%&63.2\%&62.8\%&.980&.664\\
      STTM-L(Sub1, \uppercase\expandafter{\romannumeral 2}) &.422&.490&\bf{89.1}\%&\bf{89.0}\%&64.8\%&63.6\%&.974&.664\\
      STTM-L(Sub1, \uppercase\expandafter{\romannumeral 3})&\bf{.445}&\bf{.498}&87.4\%&88.6\%&\bf{67.8\%}&\bf{68.2\%}&\bf{.964}&\bf{.644}\\
      \midrule
      STTM-H(Sub1, \uppercase\expandafter{\romannumeral 1})&.150&.266&83.6\%&92.7\%&91.4\%&91.2\%&.710&.392\\
       \rowcolor{gray!20}
       STTM-H(Sub1, \uppercase\expandafter{\romannumeral 3})&\bf{.221}&\bf{.278}&\bf{92.6\%}&\bf{98.3\%}&\bf{96.0\%}&\bf{96.3\%}&\bf{.599}&\bf{.332}\\
      \bottomrule
      \end{tabular}}
       \label{nsd_sub1}
       \vspace{-1em}
\end{table}

\subsection{Further Discussions}
\subsubsection{Rationality of Training with Cross-Subject fMRI} To further validate the effects of cross-subject fMRI training, we conduct an ablation study on the data of subject 1 in NSD. Three types of experiments are conducted:  Type I means both STTM-L and STTM-H are trained with single-subject data; Type II means the STTM-L is trained with single-subject data but the high-level guidance is from STTM-H trained with cross-subject data; Type III means STTM-H and STTM-L are trained with cross-subject data. The results are presented in \cref{nsd_sub1}.  As we can see, both STTM-L and STTM-H benefit from cross-subject fMRI training.
We can also conduct a qualitative comparison with MindEye. When not guided by high-level perception, our STTM-L model shares the same architecture as MindEye's low-level model but is trained with cross-subject fMRI data using subject adapters. We only utilize L1 loss, removing the auxiliary contrastive loss used in MindEye for low-level pipeline training. As depicted in \cref{nsd_gen_compare}, STTM-L (w/o guidance) outperforms MindEye (Low-level) across all metrics despite employing a simpler loss function. Similar performance enhancements are observed with our STTM-H (w/o GVLC) model, which has the same MLP backbone and training loss functions as MindEye but surpasses it on retrieval tasks and high-level reconstruction metrics. A potential explanation is that training with cross-subject fMRI data enhances model generalizability by exposing the shared decoding model to a more diverse data distribution and extracting more essential fMRI features. From a parameter efficiency perspective, our STTM framework trains a single model for multiple subjects, whereas MindEye trains subject-specific models with significantly more parameters. Considering these observations, training future decoding models with cross-subject data appears to be a reasonable choice.

\subsubsection{Influence of Global Visual-Linguistic Contrastive Learning}
Combining results from \cref{nsd retrival} and \cref{nsd_gen_compare}, we observe that employing the global visual-linguistic contrastive learning significantly improves text retrieval performance while minimally influencing image and brain activity retrieval tasks. It also positively impacts image reconstruction, aligning with observations from BrainCLIP despite differences in model architectures. These results contribute to understanding the multi-sensory nature of the human brain and offer a foundation for future advancements in generic multi-modal brain decoding.

\subsubsection{Influence of the High-Level and Low-Level Perception Interaction}

As shown in \cref{nsd_gen_compare}, integrating high-level perception guidance improves our low-level pipeline performance across almost all metrics. Surprisingly, even low-level metrics benefit from high-level information. Moreover, combining both pipelines enhances final reconstruction performance compared to using only the high-level pipeline. Our overall framework effectively mimics the interaction between bottom-up and top-down processes in the human brain, resulting in notable performance gains. More results are available in the appendix.

\section{Conclusion}
In this work, we introduce STTM, a novel method leveraging cross-subject fMRI data to learn transferable neural representations shared across human brains. We pre-train our models on 4 subjects from the NSD dataset and perform transfer learning on the GOD dataset. Our method achieves comparable or better decoding performance compared to previous state-of-the-art works across various tasks, including image retrieval, text retrieval, brain imaging retrieval, and image reconstruction. Notably, compared to Mind-Vis and fMRI-PTE, two pertaining-based fMRI decoding models, our method demonstrates superior reconstruction performance, suggesting its potential as an alternative approach for training fMRI foundation models. We also propose a pixel-wise reconstruction pipeline guided by high-level perceptions, showing the benefits of incorporating high-level information into pixel-wise reconstruction. Our complete STTM model comprises a high-level pipeline and a low-level pipeline, providing insights into the interaction of bottom-up and top-down processes in the human brain. However, our method has its limitations. For instance, during the pre-training stage, each subject requires a unique adapter, limiting the total number of subjects involved in pre-training due to GPU memory restrictions. Thus, subjects with more training samples are preferred.

\section*{Acknowledgements}
This work was supported by the National Natural Science Foundation of China (NO. 62088102, NO.62076235), STI2030-Major Projects (NO. 2022ZD0208801), and China National Postdoctoral Program for Innovative Talents from China Postdoctoral Science Foundation (NO. BX2021239).

%
%
\bibliographystyle{splncs04}
\bibliography{main}
\clearpage
\section*{Appendix}
\setcounter{section}{0} 
\setcounter{table}{0}   
\setcounter{figure}{0}
\renewcommand{\thetable}{A\arabic{table}}
\renewcommand{\thefigure}{A\arabic{figure}}
\begin{figure*}
\begin{center}
   \includegraphics[width=1.0\linewidth]{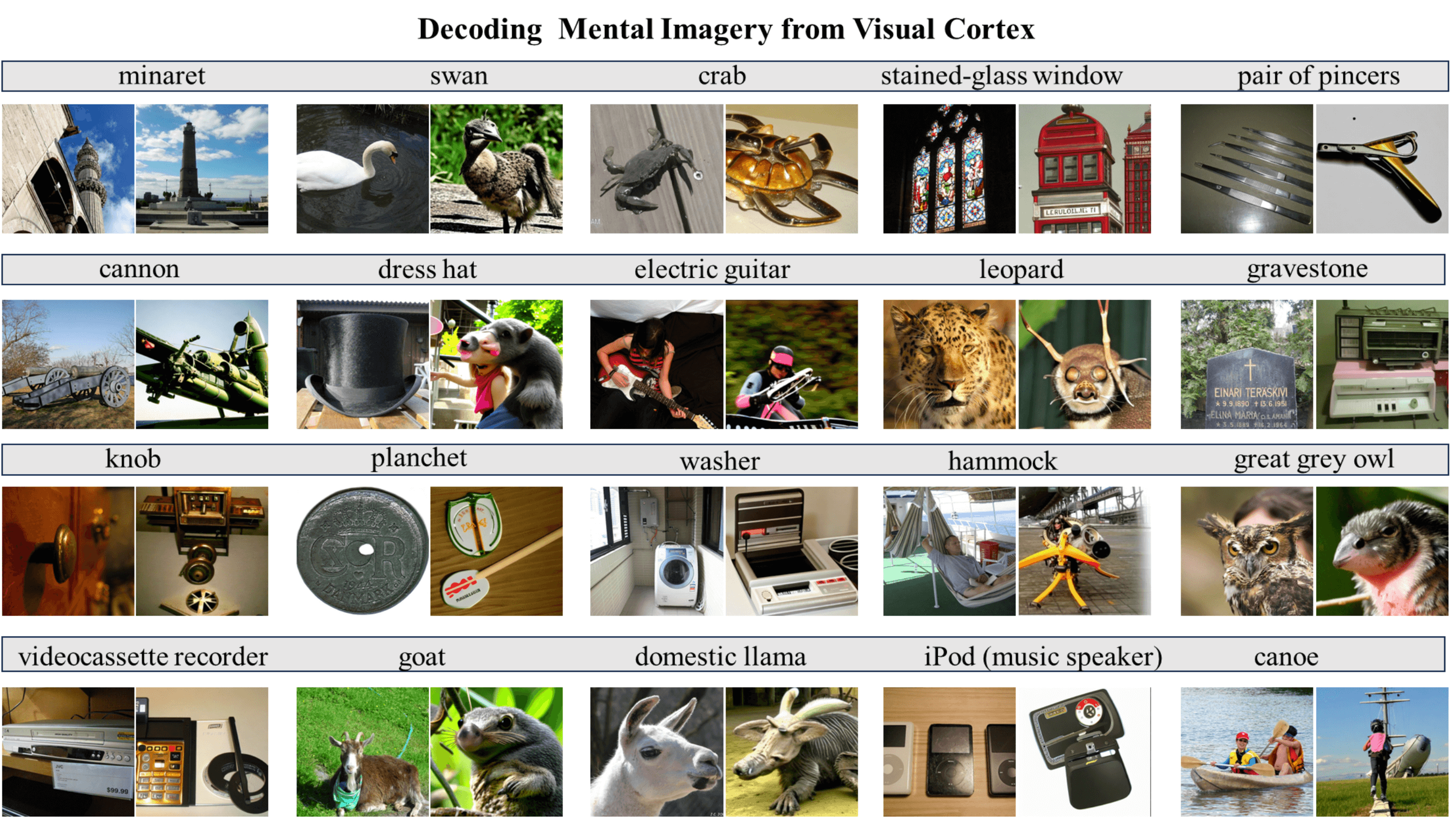}
\end{center}
   \caption{Utilizing the imagery experiment data provided in the Generic Object Decoding (GOD) dataset\cite{horikawa2017generic}, we try to visualize mental images from brain activities in the visual cortex. These brain activities were recorded while subjects freely imagined an object prompted by text cues, with their eyes closed, thus lacking ground truth images for comparison. Importantly, the object categories used in this experiment do not overlap with the GOD training set. To generate these visualizations, we averaged the fMRI patterns across 10 trials for each category. We present each category name alongside a reference image (on the left) and a generated image (on the right) to illustrate the decoded mental imagery. These results, obtained with subject 3, utilize exactly the same STTM-H model as detailed in the main body of this paper. }
\label{fig:imagine_decoding}
\end{figure*}


\subsection{Decoding Mental Imagery from Visual Cortex}
\label{imagery}
Mental imagery, the ability to create mental images without external stimuli, often produces weaker or less vivid images from memory compared to those evoked by sensory input, yet both types rely on the visual system \cite{dijkstra2018differential}. In line with this, contemporary theories of mental imagery suggest shared mechanisms between human vision and mental imagery, which implies that both perceptual visual processing and internally generated imagery activate extensive, overlapping networks of brain regions.\cite{ganis2004brain,kosslyn1999role,lee2012disentangling,slotnick2005visual}. 

Building upon prior research and utilizing the imagery experiment data collected by Horikawa and Kamitani\cite{horikawa2017generic}, we endeavor to visualize mental imagery directly from the visual cortex. Note that these data were gathered while subjects were freely imagining objects cued by text, with their eyes closed. Thus, there are no ground-truth images. We display our results for subject 3 in \cref{fig:imagine_decoding}, where we provide each category name with a reference image(on the left) and a generated image(on the right).  We also provide the results for the 50-way classification task in \cref{god_each_sub_imagine} to provide a benchmark for future research. All results are derived using the same STTM-H models as detailed in the main body of this paper and trained with the GOD training set.

As we can see, the decoders trained on brain activity elicited by visual stimuli demonstrate the capacity to generalize to decode imagery-induced brain activity. However, their performance is observed to be lower in accuracy when compared to decoding from stimulus-induced brain activity. This suggests that while there is some transferability between decoding from stimulus-induced brain activity and imagery-induced brain activity, the latter presents unique challenges that impact decoding accuracy.

\begin{table}[t]
  \caption{Quantitative results of decoding the categories of the imagined objects. There are 50 classes in total, thus the chance levels for top-1 and top-5 accuracy are 2.0\% and 10.0\%, respectively. It's important to note that there are no previous works that report such results on the GOD dataset, and we aim to provide a benchmark for future research in this area.}
    \centering
    \begin{tabular}{ccccc}
        \toprule
        Methods&{Modality}&{Prompt}&top-1 &top-5\\
        \midrule
        STTM-H(Sub1)&V\&T&Text&6.&16.0\\
        STTM-H(Sub2)&V\&T&Text&10.0&30.0\\
        STTM-H(Sub3)&V\&T&Text&12.0&36.0\\
       STTM-H(Sub4)&V\&T&Text&4.0&26.0\\
      STTM-H(Sub5)&V\&T&Text&2.0&18.0\\  
      \bottomrule
      \end{tabular}
       \label{god_each_sub_imagine}
\end{table}

\subsection{Try Different Designs for Pixel-Wise Reconstruction}
In our paper's main body, we conducted a comparative analysis of our low-level pipeline's performance under two conditions: with and without high-level perception guidance. Here, we present another design for the low-level pipeline, wherein it shares the same adapters with the pre-trained high-level pipeline and is trained without high-level perception guidance. The results are depicted in \cref{compare_adapters}. It is evident from the results that features from the high-level adapters yield poorer performance on two pixel-wise metrics (i.e., PixCorr and SSIM), which means that they contain less pixel-wise information. Our final design, which employs low-level adapters with high-level guidance, effectively combines the advantages of the other two designs.

\begin{table*}[t]
  \caption{Test different low-level features for the pixel-wise reconstruction pipeline. The low-level adapter(w/o guidance) means the low-level pipeline uses the features processed by the low-level adapters and is trained without high-level perception guidance. The high-level adapter(w/o guidance) means the low-level pipeline shares the adapters with the pre-trained high-level pipeline and is also trained without guidance. As we can see, our final design of using low-level adapters with high-level guidance combines the advantages of the first two designs. }
\resizebox{\linewidth}{!}{
    \centering
    \begin{tabular}{ccccccccc}
        \toprule
        \multirow{2}*{Low-level feature }&\multicolumn{4}{c}{Low- Level}&\multicolumn{4}{c}{High-Level} \\
        \cmidrule(r){2-5} \cmidrule(r){6-9}\\
        &PixCorr$\uparrow$&SSIM$\uparrow$&Alex(2)$\uparrow$&Alex(5)$\uparrow$&Incep$\uparrow$& CLIP$\uparrow$&Eff$\downarrow$&SwAV$\downarrow$\\
        \midrule
        Low-level adapter(w/o guidance)&.372&.488&79.6\%&79.6\%&63.6\%&63.0\%&.985&\bf{.643}\\
         High-level adapter(w/o guidance)&.285&.439&\bf{91.6\%}&\bf{87.7\%}&\bf{68.6\%}&61.6\%&\bf{.946}&.735\\
         Low-level adapter(with guidance)&\bf{.383}&\bf{.488}&83.3\%&86.0\%&68.2\%&\bf{67.1}\%&.968&.647\\
      \bottomrule
      \end{tabular}}
       \label{compare_adapters}
\end{table*}

\subsection{Ablation Study on Img2Img Strength}
\label{img2img}
The Img2Img strength controls how much information of the initial image can be maintained in the translated image. We test with different Img2Img strength values for image reconstruction on NSD. The results are reported in \cref{img2img_strength}.  As the img2img strength increases, the results for low-level metrics will increase but the results for the high-level metrics will first increase and then decrease.  The final values used in the main body of this paper are 0.3 for NSD and 0.2 for GOD.

\begin{table*}[t]
 \caption{Ablation study about the influence of the img2img strength for image reconstruction on NSD. As the img2img strength increases, the results for low-level metrics will increase but the results for the high-level metrics will first increase and then decrease. All results are averaged across 4 subjects}
    \centering
    \begin{tabular}{ccccccccc}
        \toprule
        \multirow{2}*{Img2Img Strength}&\multicolumn{4}{c}{Low- Level}&\multicolumn{4}{c}{High-Level} \\
        \cmidrule(r){2-5} \cmidrule(r){6-9}\\
        &PixCorr$\uparrow$&SSIM$\uparrow$&Alex(2)$\uparrow$&Alex(5)$\uparrow$&Incep$\uparrow$& CLIP$\uparrow$&Eff$\downarrow$&SwAV$\downarrow$\\
        \midrule
    0.0(only high-level)&.209&.276&91.5\%&97.8\%&95.4\%&95.6\%&.612&.344\\
      0.15&.316&.309&95.3\%&98.5\%&95.7\%&95.7\%&.608&.338\\
       0.2&.322&.315&95.4\%&98.4\%&95.7\%&\bf{95.9\%}&.608&\bf{.337}\\
   0.25&.327&.324&95.4\%&98.5\%&95.7\%&95.8\%&\bf{.608}&.338\\
   $0.3^*$&.333&.334&\bf{95.7\%}&\bf{98.5\%}&\bf{95.8\%}&95.7\%&.611&.338\\
   1.0(only low-level)&\bf{.383}&\bf{.488}&83.3\%&86.0\%&68.2\%&67.1\%&.968&.647\\
      \bottomrule
      \end{tabular}
       \label{img2img_strength}
\end{table*}

 \subsection{Subject-Specific Results On NSD}
 \label{nsd_individual}
 In \cref{nsd_each_sub}, we provide the image reconstruction results and retrieval results for each subject and each pipeline on the NSD dataset. 
\begin{table*}
 \caption{Image reconstruction results and retrieval results for each subject and each pipeline on the NSD dataset}
\resizebox{\linewidth}{!}{
    \centering
    \begin{tabular}{cccccccccccc}
        \toprule
        \multirow{2}*{Methods}&\multicolumn{4}{c}{Low- Level}&\multicolumn{4}{c}{High-Level}&\multicolumn{3}{c}{Retrieval tasks} \\
        \cmidrule(r){2-5} \cmidrule(r){6-9}\cmidrule(r){10-12}\\
        &PixCorr$\uparrow$&SSIM$\uparrow$&Alex(2)$\uparrow$&Alex(5)$\uparrow$&Incep$\uparrow$& CLIP$\uparrow$&Eff$\downarrow$&SwAV$\downarrow$&Image@1$\uparrow$&Text@5$\uparrow$&Brain@1$\uparrow$\\
        \midrule
        STTM-H(Sub1)&.221&.278&92.6\%&98.3\%&96.0\%&96.3\%&.599&.332&97.0\%&43.9\%&98.4\%\\
        STTM-H(Sub2)&.211&.275&92.0\%&97.9\%&95.2\%&95.3\%&.619&.346&96.5\%&38.9\%&97.8\%\\
        STTM-H(Sub5)&.197&.275&91.5\%&98.0\%&96.2\%&96.4\%&.596&.337&89.5\%&46.3\%&92.4\%\\
        STTM-H(Sub7)&.205&.274&89.7\%&96.6\%&94.1\%&94.5\%&.635&.360&88.1\%&36.0\%&90.9\%\\
        \midrule
        STTM-L(Sub1)&.445&.498&87.4\%&88.6\%&67.8\%&68.2\%&.964&.644&-&-&-\\
        STTM-L(Sub2)&.396&.490&85.5\%&87.9\%&68.1\%&67.7\%&.964&.645&-&-&-\\
        STTM-L(Sub5)&.351&.485&80.6\%&84.5\%&69.4\%&67.2\%&.969&.649&-&-&-\\
        STTM-L(Sub7)&.338&.481&79.8\%&83.0\%&67.2\%&65.4\%&.974&.653&-&-&-\\
        \midrule
         STTM(Sub1)&.390&.343&97.5\%&99.1\%&96.5\%&96.5\%&.595&.325&97.0\%&43.9\%&98.4\%\\
         STTM(Sub2)&.343&.337&96.4\%&98.8\%&96.0\%&95.5\%&.619&.340&96.5\%&38.9\%&97.8\%\\
         STTM(Sub5)&.305&.330&95.0\%&98.4\%&96.9\%&96.6\%&.594&.331&89.5\%&46.3\%&92.4\%\\
         STTM(Sub7)&.293&.326&93.9\%&97.7\%&93.9\%&94.3\%&.635&.356&88.1\%&36.0\%&90.9\%\\
      \bottomrule
      \end{tabular}}
       \label{nsd_each_sub}
\end{table*}

\subsection{Subject-Specific Results On GOD}
\cref{god_each_sub} shows the individual results for the  zero-shot classification task  on the test set of the GOD dataset
\begin{table}
     \caption{Zero-shot classification results for each subject on the test set of the GOD dataset }
    \centering
    \begin{tabular}{ccccc}
        \toprule
        \multirow{2}*{Methods}& \multirow{2}*{Modality}&\multirow{2}*{Prompt}&\multicolumn{2}{c}{Average}\\
        \cmidrule(r){4-5}
        &&&top-1 &top-5\\
        \midrule
        STTM-H(Sub1)&V\&T&Text&18.0&60.0\\
        STTM-H(Sub2)&V\&T&Text&20.0&58.0\\
        STTM-H(Sub3)&V\&T&Text&26.0&68.0\\
        STTM-H(Sub4)&V\&T&Text&28.0&62.0\\
        STTM-H(Sub5)&V\&T&Text&24.0&62.0\\
        
      \bottomrule
      \end{tabular}
       \label{god_each_sub}
\end{table}

\subsection{Contrastive Losses in This Work }
\label{loss}
\subsubsection{BiMixCo Loss}
In BiMixCo\cite{scotti2023reconstructing} loss, voxels are mixed using a factor $\lambda$ sampled from the Beta distribution
with $\alpha = \beta = 0.15$.

\begin{align}x_{mix_{i,k_i}}= \lambda_i· x_i + (1 -\lambda_i) · x_{k_i}, p_i^*=f(x_{mix_{i,k_i}}),p_i=f(x_i), t_i=CLIP_{img}(y_i),
\end{align}
where $x_i$ and $y_i$ are the $i$-th fMRI sample and image respectively. $k_i \in [1, N]$ is an arbitrary mixing index for the $i$-th datapoint and $f$ represents the combined MLP and projector. $p^*$, $p$ and $t$ are L2-normalized. The BiMixCo is defined as:


\begin{align}
L_{BiMixCo}=-\sum_{i=1}^{N}[\lambda_{i}log(\frac{exp(\frac{p_i^*\cdot t_i}{\tau})}{\sum_{m=1}^{N}exp(\frac{p_i^*\cdot t_m}{\tau})})+(1-\lambda_i)log(\frac{exp(\frac{p_i^*\cdot t_{k_i}}{\tau})}{\sum_{m=1}^{N}exp(\frac{p_i^*\cdot t_m}{\tau})})]\\\notag
-\sum_{j=1}^{N}[\lambda_{j}log(\frac{exp(\frac{p_j^*\cdot t_j}{\tau})}{\sum_{m=1}^{N}exp(\frac{p_m^*\cdot t_j}{\tau})}))+\sum_{\{l|k_l=j\}}(1-\lambda_l)log(\frac{exp(\frac{p_l^*\cdot t_j}{\tau})}{\sum_{m=1}^{N}exp(\frac{p_m^*\cdot t_j}{\tau})})],
\end{align}
where $\tau$ is a temperature hyperparameter, and N is the
batch size.
\subsubsection{SoftCLIP Loss}
With the same notation, SoftCLIP\cite{scotti2023reconstructing} loss is defined as:
\begin{align}
\label{eq4}
L_{SoftCLIP}=-\sum_{i=1}^{N}\sum_{j=1}^{N}[\frac{exp(\frac{t_i\cdot t_j}{\tau})}{\sum_{m=1}^{N}exp(\frac{t_i\cdot t_m}{\tau})}log(\frac{exp(\frac{p_i\cdot t_i}{\tau})}{\sum_{m=1}^{N}exp(\frac{p_i\cdot t_m}{\tau})})],
\end{align}

\subsection{Reconstruction Evaluations Metrics}
\label{metrics}
\paragraph{PixCorr:}The pixel-level correlation of reconstructed and ground-truth images.
\paragraph{SSIM:} The structural similarity
index metric\cite{wang2004image}.
\paragraph{Alex(2) \& Alex(5):} The 2-way comparisons of the second and fifth layers of AlexNet\cite{krizhevsky2012imagenet}, respectively.
\paragraph{Inception("Incep"):} The 2-way comparison of the last pooling layer of Inception-V3\cite{szegedy2016rethinking}. 
\paragraph{CLIP:} The 2-way comparison of the output layer
of the CLIP-Vision\cite{radford2021learning} model.

\paragraph{EffNet-B1(“Eff”):} Distance metrics gathered from EfficientNet-B1\cite{tan2019efficientnet}. 
\paragraph{SwAV:} Distance metric gathered from SwAV-ResNet50\cite{caron2020unsupervised} model. 

Two-way identification followed the methods of Ozcelik and VanRullen\cite{ozcelik2023natural} and Scotti et al.\cite{scotti2023reconstructing}. We computed Pearson correlations between embeddings for the ground truth image and the reconstructed image, as well as between the ground truth image and another reconstruction from the test set. Correct identification occurred when the correlation with the ground truth image was higher than with the other reconstruction. Performance for each test sample was averaged across all possible pairwise comparisons using the remaining 981 reconstructions to prevent bias from random sample selection.


\subsection{More Reconstruction Examples}
\begin{figure*}
\begin{center}
   \includegraphics[width=1.0\linewidth]{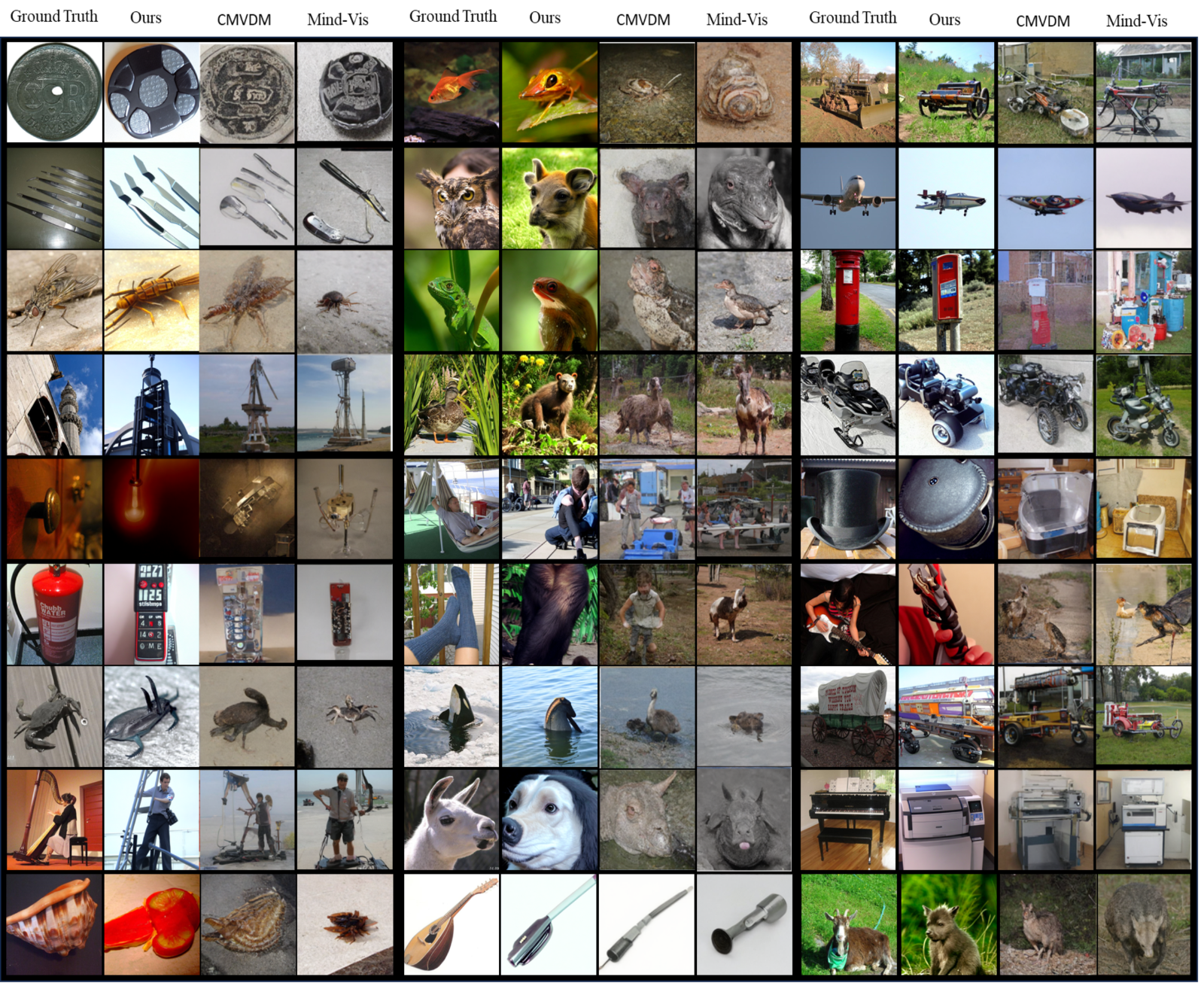}
\end{center}
   \caption{Qualitative comparison with state-of-the-art methods on the GOD dataset. From the intuitive point of view, our method gets similar or even better results compared to CMVDM\cite{zeng2023controllable} and Mind-Vis\cite{chen2023seeing}.}
\label{fig:GOD_SOTA_img}
\end{figure*}

In \cref{fig:GOD_SOTA_img}, we qualitatively compare our reconstructions against those of previous state-of-the-art works. From the intuitive point of view, our method gets similar or even better results compared to CMVDM\cite{zeng2023controllable} and Mind-Vis\cite{chen2023seeing}.

\begin{figure*}[t]
\begin{center}
   \includegraphics[width=1.0\linewidth]{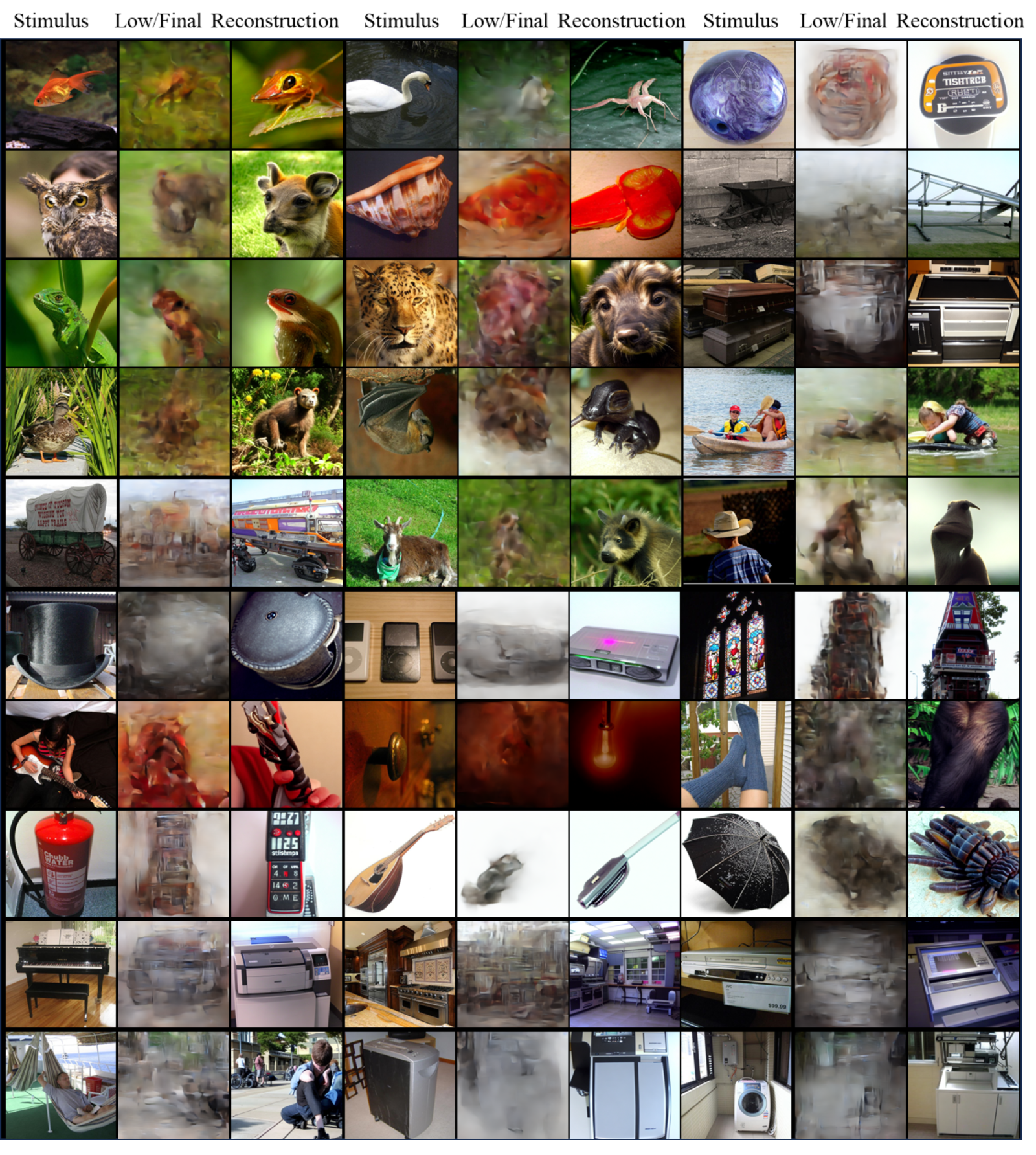}
\end{center}
   \caption{More reconstruction samples for subject 3 from GOD}
\label{fig:more_sample_GOD}
\end{figure*}

In \cref{fig:more_sample_GOD}, we give more reconstruction samples for subject 3 of the GOD dataset.

\begin{figure*}[t]
\begin{center}
   \includegraphics[width=1.0\linewidth]{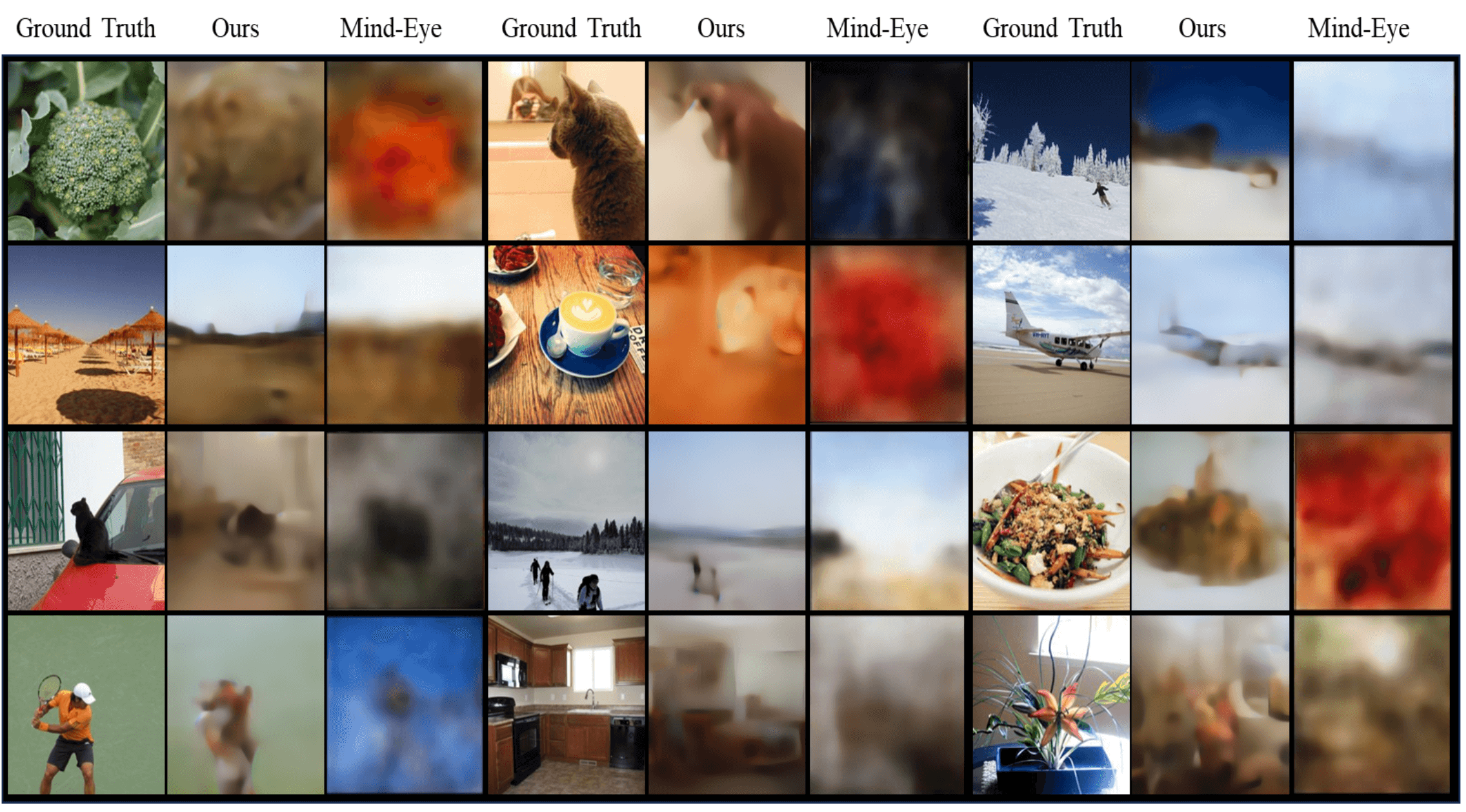}
\end{center}
   \caption{Qualitative comparison with MindEye for low-level reconstruction pipeline. Our reconstructions contain more pixel-level information like boundary and color distribution.}
\label{fig:low_level_compare}
\end{figure*}

In \cref{fig:low_level_compare}, we compare our low-level reconstructions with those of MindEye\cite{scotti2023reconstructing}. Our reconstructions contain more details and are more faithful to the original color distribution.

\begin{figure*}[t]
\begin{center}
   \includegraphics[width=1.0\linewidth]{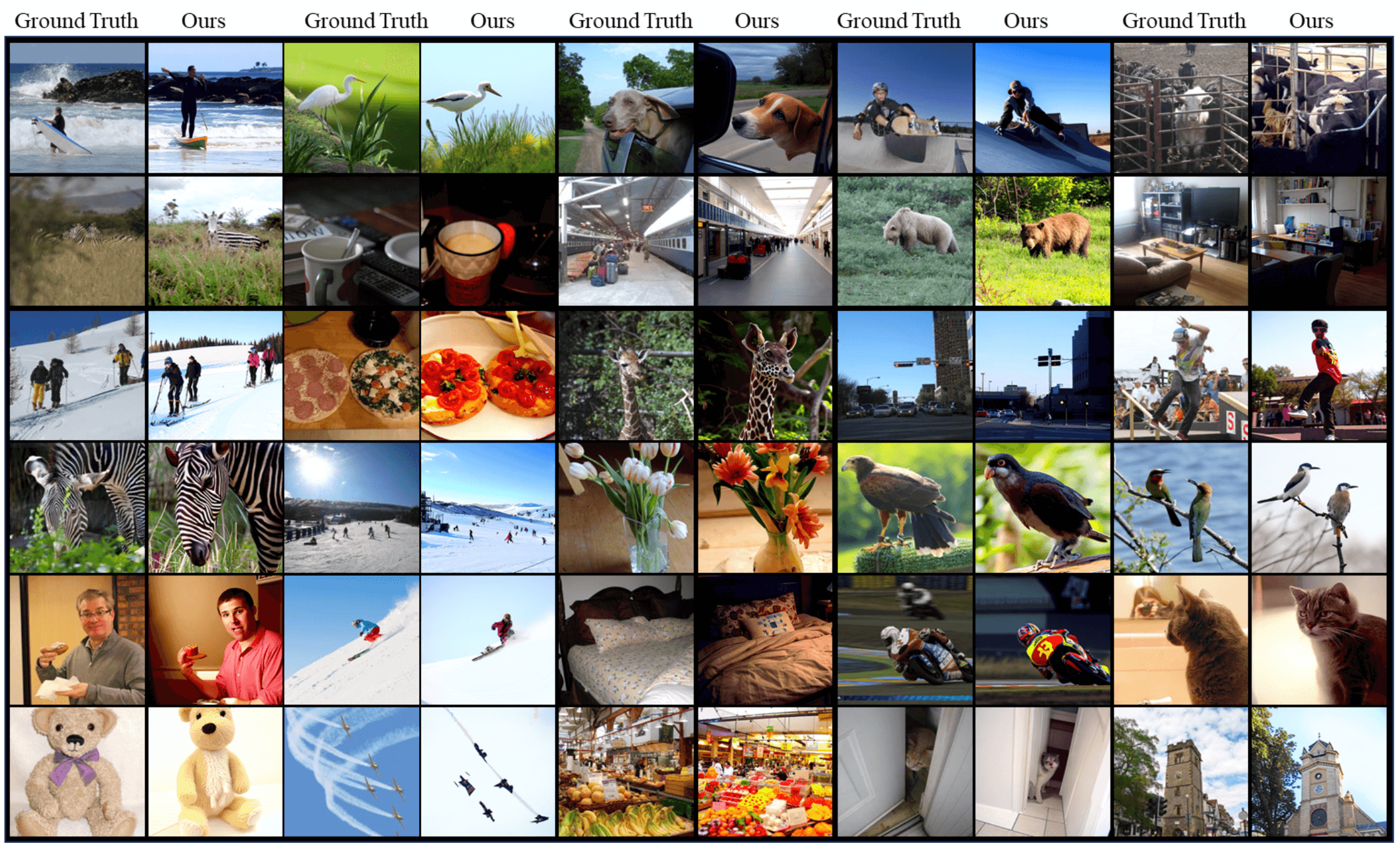}
\end{center}
   \caption{More reconstruction samples for subject 1 from NSD}
\label{fig:more_samples_NSD}
\end{figure*}

In \cref{fig:more_samples_NSD}, more reconstructions for subject 1 from the NSD dataset are visualized.

\subsection{Prompts for classification}

The text prompts used in this work are the same as BrainCLIP\cite{liu2023brainclip}:

"a photo of a \{\}.",

        "a blurry photo of a \{\}.",
        
        "a black and white photo of a \{\}.",
        
        "a low contrast photo of a  \{\}.",
        
        "a high contrast photo of a  \{\}.",
        
        "a bad photo of a  \{\}.",
        
        "a good photo of a  \{\}.",
        
        "a photo of a small \{\}.",
        
        "a photo of a big  \{\}.",
        
        "a photo of the  \{\}.",
        
        "a blurry photo of the  \{\}.",
        
        "a black and white photo of the  \{\}.",
        
        "a low contrast photo of the  \{\}.",
        
        "a high contrast photo of the  \{\}.",
        
        "a bad photo of the  \{\}.",
        
        "a good photo of the  \{\}.",
        
        "a photo of the small  \{\}.",
        
        "a photo of the big \{\}."   
        
The “\{\}” is replaced by a specific class name. Their embeddings are averaged to get the classification weights.
\subsection{Potential Social Impact}
 Research in AI for brain decoding holds immense potential for various social impacts. One significant area of impact lies in healthcare, where advancements could revolutionize diagnosis and treatment for neurological disorders, such as Alzheimer's, Parkinson's, and epilepsy. Accurate decoding of brain signals could lead to early detection of cognitive decline, enabling timely interventions and personalized treatment plans. Moreover, the development of brain-computer interfaces (BCIs) could greatly enhance communication and mobility for individuals with severe disabilities, empowering them to interact with the world more independently. Additionally, ethical considerations surrounding privacy, consent, and data security must be carefully addressed to ensure the responsible and equitable deployment of such technology.  In this work, all the used data are obtained from public projects and used responsibly.

\end{document}